# Segmentation of Breast Regions in Mammogram Based on Density: A Review

Nafiza Saidin[1], Harsa Amylia Mat Sakim[1], Umi Kalthum Ngah[1] and Ibrahim Lutfi Shuaib[2]

[1] Imaging & Computational Intelligence Group (ICI)
School of Electrical and Electronic Engineering
Universiti Sains Malaysia, Engineering Campus,
14300 Nibong Tebal, Seberang Perai Selatan,
Pulau Pinang, Malaysia
*Email: ns.ld09@student.usm.my, amylia@eng.usm.my, eeumi@eng.usm.my*

[2] Advanced Medical and Dental Institute,
Universiti Sains Malaysia No 1-8 (Lot 8),
Persiaran Seksyen 4/1, Bandar Putra Bertam,
13200 Kepala Batas Pulau Pinang, Malaysia
*Email: ibrahim@amdi.usm.edu.my*

**Abstract**

The focus of this paper is to review approaches for segmentation of breast regions in mammograms according to breast density. Studies based on density have been undertaken because of the relationship between breast cancer and density. Breast cancer usually occurs in the fibroglandular area of breast tissue, which appears bright on mammograms and is described as breast density. Most of the studies are focused on the classification methods for glandular tissue detection. Others highlighted on the segmentation methods for fibroglandular tissue, while few researchers performed segmentation of the breast anatomical regions based on density. There have also been works on the segmentation of other specific parts of breast regions such as either detection of nipple position, skin-air interface or pectoral muscles. The problems on the evaluation performance of the segmentation results in relation to ground truth are also discussed in this paper.

## 1. Introduction

Breast cancer is the most prevalent cancer and is the leading terminal illness among women worldwide. Early detection of breast cancer is crutial and for that, mammography plays the most essential role as a diagnostic tool. Breast cancer usually occurs in the fibroglandular area of breast tissue. Fibroglandular tissue attenuates x-rays greater than fatty tissue making it appear bright on mammograms. This appearance is described as 'mammographic density' or also known as breast density [1]. The breast density portion contains ducts, lobular elements and fibrous connective tissue of the breast. Breast density is an important factor in the interpretation of a mammogram. The proportion of fatty and fibroglandular tissue of the breast region is evaluated by the radiologist in the interpretation of mammographic images. The result is subjective and varies from one radiologist to another.

In the study conducted by Martin *et al.* [2], hormone therapies, including estrogen and tamoxifen treatments have been found to be able to change mammographic density [3-6] and alter the risk of breast cancer [7-10]. Therefore, a method for measuring breast density can provide as a tool for investigating breast cancer risk. Subsequently, the association of breast density with the risk of breast cancer can be more definitive and will allow better monitoring response of a patient as preventive or interventional treatment of breast cancers.

Breast cancer is the leading cause of death for women in their 40s in the United States [11]. In developing Asian countries, most breast cancer patients are younger than those in developed Asian and Western countries [12, 13]. Younger patients mean that the mammographic images would be denser [14]. In a dense breast, the sensitivity of mammography for early detection of breast cancer is reduced. This may be due to the tell tale signs being embedded in dense tissue, which have similar x-ray attenuation properties. Although the incidence of breast cancer is lower in developing Asian countries, the mortality rate is higher when compared with other nations worldwide. In fact, it is the leading cause of cancer deaths in Asia and is the commonest female malignancy in developing Asian Countries [15]. Therefore, it is most appropriate to focus on density based research of mammograms especially amongst Asian women, involving younger aged patients having denser breast and thus are difficult to diagnose.

## 2. Segmentation of Breast Regions in Mammogram based on Density

Image segmentation means separating the image into similar constituent parts, including identifying and partitioning regions of interests. Segmentation is an important role and also the first vital step in image processing, which must be successfully taken before subsequent tasks such as feature extraction and classification step. This technique is important in breast applications such as localizing suspicious regions, providing objective quantitative assessment and monitoring of the onset and progression of breast diseases, as well as analysis of anatomical structures. Many researchers had focused on image processing, including segmentation technique to identify masses and calcifications in order to detect early breast cancer. Most of the image processing techniques are implemented on the whole mammogram without taking into consideration that mammograms have different density patterns and that anatomical regions are used by radiologists in the interpretation [16]. The medical community has realized breast tissue density as an important risk indicator for the growth of breast cancer [17- 21]. Wolfe has noticed that the risk for breast cancer growth is determined by mammography parenchymal patterns [22], and it has also been confirmed by other researchers, such as Boyd *et al.* [23], van Gils *et al.* [24] and Karssemeijer [25]. Before classification or segmentation is performed, a proper understanding of breast anatomical regions is essential.

2.1 Mammogram and Breast Regions

A mammogram is an x-ray projection of the 3D structures of the breast. It is obtained by compressing the breast between two plates. Mammograms have an inherent "fuzzy" or diffuse appearance compared with other x-rays or Computed Tomography images. This is due to the superimposition of densities from differing breast tissues, and the differential x-ray attenuation characteristics associated with these various tissues. A mammogram contains two different regions: the exposed breast region and the unexposed air-background (non-breast) region. Background region in a mammogram usually appears as a black region, and it also contains high intensity parts such as bright rectangular labels, opaque markers, and artifacts (e.g. scratches). Breast regions can be partitioned into:
1. Near-skin tissue region, which contains uncompressed fatty tissue, positioned at the periphery of the breast, close to the skin-air interface where the breast is poorly compressed.
2. Fatty region, which is composed of fatty tissue that is positioned next to the uncompressed fatty tissues surrounding the denser region of fibroglandular tissue.
3. Glandular regions, which are composed of non uniform breast density tissue with heterogeneous texture that surrounds the hyperdense region of the fibroglandular tissue.
4. Hyperdense region, which is represented by high density portions of the fibroglandular tissue, or can be a tumor.

Fig. 1 shows a mammogram image, with different breast tissues and Fig 2 demonstrates the illustration of different breast regions when the breast tapers off. The breast boundary can be obtained by partitioning the mammogram into breast and background regions. The extracted breast boundary should adequately model the skin-air interface and preserve the nipple in profile. However, skin-line region in mammograms where the breast tapers off is normally very low in grey-level contrast. It is caused by the lack of uniform compression of the breast, near the breast edge region [26]. This effect decreases the visibility along the peripheral region of the mammogram and makes it difficult to preserve the breast skin-line and to identify the nipple position as shown in Fig. 2.

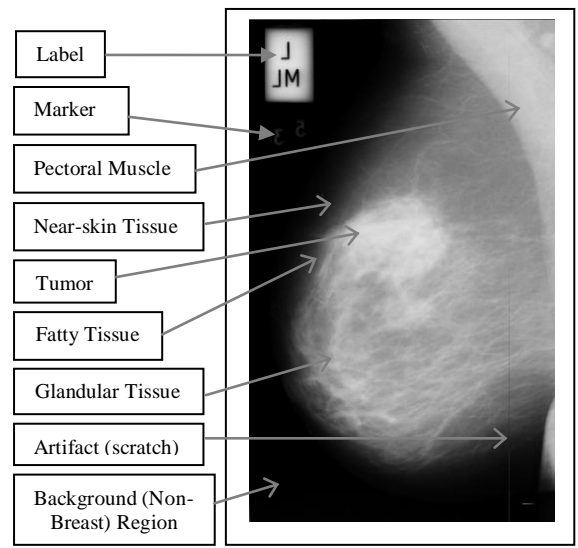

Fig. 1 A mammogram image composes of the image background, label, marker, artifact (scratch), near-skin tissue, fatty tissue, pectoral muscle and denser glandular tissue.

Breast density is a measurement of the dense structure of fibroglandular tissue, which appears white on a mammogram. Fibroglandular tissues appear to have disc or cone shapes and extend through the interior of the breast from the region near the chest wall to the nipple [27]. The breast density part contains ducts, lobular elements, and fibrous connective tissue of the breast. Fatty tissues are less dense and appear as darker regions. So, if the tumour is in the fatty region, it is easier to be interpreted compared to if it is in the fibroglandular region. According to Caulkin *et al.* [28], in clinical practice, they realized that the majority of cancers are associated with glandular rather than fatty tissues. Tumors generally appear similar to hyperdense parts compared to their surroundings tissues. The density of dense structures such as the milk ducts is similar to the tumor making it difficult to interpret. It is tedious to differentiate between normal, dense tissue and cancerous tissue when the tumor is surrounded by glandular tissues [14]. So, in order to clarify these regions, segmentation techniques should be adapted. It is important to detect the glandular tissue and highlight the hyperdense part of glandular tissue that possibly contains a tumor. It is difficult to compare the two regions having similar intensities using the naked eyes, but it is possible to do this using computer-aided detection through segmentation.

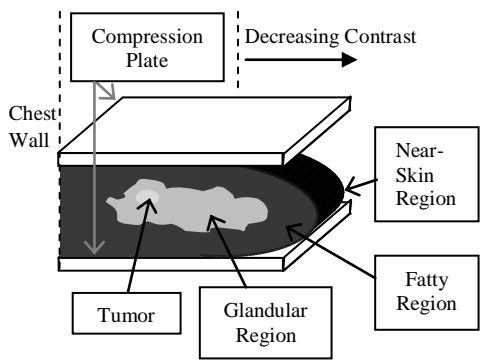

Fig. 2 Illustration of different breast regions when the breast tapers off.

Wolfe categorized breast density into four patterns. Quantitative classification of breast density into six categories has been developed by Byng *et al.* [29] and Boyd *et al.* [23]. According to Byng *et al.* [29], in the quantification, it is difficult to evaluate a volume of dense tissue because it is highly dependent on the compressed thickness during the mammographic examination and also on the spectrum of the x-ray beams. Optionally, the proportion of the breast area representing dense tissue is used for the quantification of mammographic density. Byng *et al.* [29] performed segmentation using an interactive thresholding technique of the dense tissue. Quantification is then obtained automatically by counting pixels within the regions recognized as the dense tissue. The research provides benefits in the risk assessment of breast cancers and also for monitoring changes in the breast density as prevention procedures. The segmentation using thresholding technique in the study by Byng *et al.* [29] is limited to the cranio-caudial view of the mammogram image. However, for the media-lateral oblique view, the study suggested the option of suppressing the pectoral muscle. Breast Imaging Reporting and Data System (BIRADs), which was developed by the American College of Radiology (ACR) is the recent standard in radiology for categorizing the breast density [30]. BIRADs classify breast density into four major categories: (1) predominantly fat; (2) fat with some fibroglandular tissue; (3) heterogeneously dense; and (4) extremely dense. According to Zhou *et al.* [31], there is a large inter-observer variability in providing BI-RADS ratings among experienced radiologists. They suggested an automatic and quantitative method for breast density estimation, which is reproducible and can reduce inter and intra-observer variabilities.

2.2 Segmentation of Fibroglandular Tissue

According to Suckling *et al.* [33], automated segmentation of glandular tissue or parenchymal pattern can provide as the beginning step in mammographic lesion detection. Segmentation of abnormal structures in the breast, consequently, depends on breast tissue density. Segmentation of the glandular tissue can also supply as a primary step in order to detect the suspicious mass and to reduce false positives. Usually, mass is represented by hyperdense structure. Overlapped fibroglandular tissue also has similar intensity with mass [16]. Hence, by focusing on glandular area and highlighting the hyperdense regions of the glandular area, it can assist and contribute as a second opinion for experts in diagnosis. According to Miller & Astley [33], identification of glandular tissue in a mammogram is necessary for assessing asymmetry between the left and right breasts. According to Matsubara *et al.* [34] the assessment of fibroglandular tissue can be used to estimate the degree of risk that the lesions are obscured by normal breast tissue and also to suggest another examination such as breast ultrasound. The combination of mammogram and ultrasound is effective in depicting breast cancer. Therefore, there is a need to develop a system, which can segment the glandular tissue area automatically.

Ferrari *et al.* [35] segmented the fibroglandular disc with a statistical method based on a Gaussian mixture modeling. Mixtures of up to four weighted Gaussians represent a particular density class in the breast. Grey-level statistics of the pectoral muscles were used to determine the tissue region that represents the fibro-glandular disc. Ols´en & Mukhdoomi [36] used Minimum Cross-Entropy to obtain an optimum threshold for detecting glandular tissue automatically. The idea of Masek [37] is used for fully automated segmentation algorithm extracting the glandular tissue disc from mammograms. Similar to Ferrari [35], El-Zaart [38] also used statistical approach for detecting the fibroglandular disc. Ferrari used Gaussian Mixture Modelling while El-Zaart used Gamma Mixture Modelling. According to El-Zaart [38], Gamma based method detected more precisely the fibro-glandular disc regions; while Gaussian based method falsely detected more regions that are not part of the glandular discs. Several other researchers had also segmented the fibroglandular discs and classified the glandular tissue into 2 to 4 categories.

2.3 Classification of Breast based on Density

There exists numerous classification research based on breast density. Miller and Astley [33] used granulometry and texture energy to classify breast tissue into fatty and glandular breast types. Taylor *et al.* [39] classified fatty and dense breast types using an automated method of extracting the Region of Interest (ROI) based on texture. Karssemeijer [25] used four categories in the classification of the density. Bovis and Singh [40] analysed two different classification methods, which are four-class categories according to the BIRADS system and two-class categories, differentiating between dense and fatty breast types. Sets of classifier outputs are combined using six different classifier combination rules proposed by Kittler *et al.* [41] and the results were compared. The results showed that the classification based on BIRADS system for the four-class categories (average recognition rate, 71.4%) is a challenging task in comparison to the two-class categories (average recognition rate, 96.7%). Zhou *et al.* classified breast density into one of four BIRADS categories according to the characteristic features of gray level histogram [31]. They found that the correlation between computer-estimated percent dense area and radiologist manual segmentation was 0.94 and 0.91 with root-mean-square (RMS) errors at 6.1% and 7.2%, respectively, for CC and MLO views. Matsubara, *et al.* [34] divided breast mammogram images into three regions using variance histogram analysis and discriminant analysis. Then, they classify it into four categories, which are (1) fatty, (2) mammary gland diffuseness, (3) non-uniform high density, and (4) high density, by using the ratios of each of the three regions. Torrent *et al.* [42] used a previously developed approach by Oliver *et al.* [43], which adopted a Bayesian combination of the C4.5 Decision tree and the k-Nearest Neighbor (kNN) algorithm to classify the breast according to BIRADS categories. Oliver *et al.* [44] implemented kNN classifier to differentiate the breast (fatty and dense).

2.4 Segmentation of Breast Anatomical Regions

Only a small group of researchers have done segmentation based on breast tissue anatomy. By doing segmentation based on the breast anatomy, more detailed divisions can be made. For example, with the detection of breast edge, distortion in breast structure and the nipple position in the breast will be detectable. This will also help in diagnosis. The segmentation method proposed by Karssemeijer [25] allowed subdivision of a mammogram into three distinct areas: breast tissue, pectoral muscle and background. For research on segmentation of breast regions into different densities, the suppression of pectoral muscle is not so significant. Instead, pectoral muscle can be used as a reference in estimating the area of glandular tissue [25, 34]. According to Karssemeijer, the density of the pectoral can be used as a reference for interpretation of densities in the breast tissue area, where regions of similar brightness with the pectoral will most likely correspond to fibro-glandular tissue. Saidin *et al.* used graph cut algorithm on mammograms to segment breast regions into the background, skin-air interface, fatty, glandular and pectoral muscle [45]. Adel *et al.* proposed segmentation of breast regions into pectoral muscle, fatty and fibroglandular regions, using a Bayesian technique with adaptation of Markov random field for detecting regions of different tissues on mammograms [46]. Aylward *et al.* segmented the breast into five regions using a combination of geometric (Gradient magnitude ridge traversal) and statistical (Gaussian mixture modeling) method [47]. The five regions that they segmented are the background, uncompressed fat, fat, dense tissue and muscle. El-Zaart segmented mammogram image into 3 regions, which are fibroglandular disc, breast region and background [38]. Most of the work done on segmentation of breast anatomical regions, automatically will detect the fibroglandular disc. However, only a handful of researchers had performed research on segmentation of fibroglandular disc and also segmentation of other breast anatomical regions.

2.5 Segmentation of Other Specific Breast Region in Breast Density Research

Most of the density based breast segmentation system involves pre-processing. Image processing technique is usually employed to detect the boundary of the breast region and to remove markers in background area of mammograms. Breast boundary detection (breast contour, breast edge, skin-air interface detection or also called skin-line estimation) is considered as an initial and essential pre-processing step. The purpose is to enable abnormality detection to be limited to the breast area without influenced from the background. By limiting the area to be processed into a specific region in an image, the accuracy and efficiency of segmentation algorithms could be increased. However, failure to detect breast skin-line accurately, could lead to the situation whereby a lesion which is located near to the breast edge may be missed [48]. Usually, research carried out on the segmentation and classifications of glandular tissue based on density would also give rise to the suppression of pectoral muscle in order to avoid incorrect segmentations. Several studies have been conducted on the suppression of pectoral muscle in the segmentation and classification of glandular tissue. In 1998, Karssemeijer proposed an automatic classification of density patterns in mammograms, including a method for automatic segmentation of the pectoral muscle in oblique mammograms, using the Hough transform. This is due to the fact that in some mammograms, the pectoral muscle has similar intensities with the glandular tissues.

Some of the work applied background and annotation subtraction to correctly focus the algorithm on the glandular tissues [44, 49]. Chatzistergos *et al.* used characteristics of monogenic signals to separate a breast region from its image background and Gabor wavelets to subtract the pectoral muscle [49]. Then, classification methods using texture characteristics [50] and probabilistic Latent Semantic Analysis (PLSA) [51] are adapted in their research. In many segmented images, the outline of the breast region is positioned more inward than the actual boundary, perhaps because the skin line is hardly visible. Segmentation research by Oliver *et al.* [43] resulted in a minor lost of skin-air regions in the breast area. Nevertheless, a few researchers have instead tried to avoid this situation by preserving the skin line or nipple position as much as possible, which in turn, helps in the architectural distortion detection [25, 45]. According to Karssemeijer, it is important to preserve the skin line position for feature selection [25].

## 3. Database of Mammograms

Several databases have commonly been used as test beds for the performance of the proposed segmentation algorithms. A large number of images are necessary to test a Computer Aided Diagnosis system and to compare processing results with others for performance evaluations. In order to overcome the difficulty in accessing hospitals and clinics confidential files, there is a need for a public database. MIAS (Mammographic Image Analysis Society Digital Mammogram Database) [52] and DDSM (Digital Database for Screening Mammography) [53] are examples of well known and broadly used mammographic databases. MIAS database is in *pgm* format with 8 bits images, and it was published in 1994. DDSM database is in LJPEG format, which is a non-standard version and needs specific libraries/software. Other examples of databases are CALMa (Computer Assisted Library for Mammography) [54], and LLNL (Lawrence Livermore National Laboratory)/UCSF database [55]. Most recently available database is LAPIMO or also known as BancoWeb LAPIMO, which can be accessed from http://lapimo.sel.eesc.usp.br/bancoweb [56]. This database emphasizes on quality of images and on variety of cases. The images are in the TIFF default format with 12 bits of contrast images, and their spatial resolutions are either 0.085 mm or 0.150 mm, depending on the scanner used. The scanners used during the digitization process are Lumiscan 50 and Lumiscan 75. These images are used to test processing techniques or segmentation algorithms developed by researchers. However, because of LAPIMO is the most recent database and it is relatively new, so very few image processing or segmentation techniques involving images from the database can be used as comparisons.

## 4. Performance Evaluation

The most essential requirement from a radiologist point of view for image processing algorithms is the ability to achieve enhanced visualizations of anatomical structure, while preserving the detail of the structure [57]. There are numerous researches, which worked on the classification and segmentation of glandular tissues. Each classification and segmentation result needs evaluation of its performance. There are three types of performance evaluations. The first type involves qualitative assessment, the second is quantitative assessment involving the ground truth evaluation, and the third is a statistical evaluation. Performance evaluation for research on classification of breast density involves comparison of research result with density class that has been given by radiologist, while performance evaluation for segmentation of breast density usually is done in qualitative analysis. This is because of the difficulty in obtaining the ground truths from radiologist. The quantitative analysis is performed only by a small number of researches. For the quantitative analysis, usually the performance of the segmentation results is compared with the ground truth by the radiologist. Ground truth in these density based research means, a correct marking of the glandular tissue or density area by the radiologist in a digital mammogram. For statistical evaluation, Receiver operating characteristic (ROC) analysis is commonly employed to ensure the validity of computer aided diagnosis systems [58]. The ROC analysis allows for a plot of the sensitivity (True Positive Fraction, TPF) against the specificity (False Positive Fraction, FPF). The area under the ROC curve ($A_z$) represents a quantitative measure of the accuracy of the segmentation or classification technique. When the value is 0, it indicates poor segmentation or classification performance while 1 indicates high segmentation or classification performance. However, it has certain restrictions and also suffers from weaknesses. Since, it is a pixel based assessment, for region based analysis, the Free Response Operating Characteristic (FROC) works better. This is equivalent to the ROC analysis, except that the false positive rate on the x-axis is replaced by the number of false positives per image. Additionally, a definition of a segmented region is required. FROC looks for location information from the result of the segmentation algorithm [59].

In the segmentation or classification based density approach, a positive case means correct detection or classification of breast glandular or dense tissue while a negative case means misclassification of other tissues as such a type. The formula and definition of the fractions are as below:
1. True Positive (TP) means breast segmented or classified as glandular/dense tissue that proved to be glandular/dense tissue.
2. False Positive (FP) means breast segmented or classified as glandular/dense tissue that proved to be other tissues.
3. False Negative (FN) means breast segmented or classified as other tissues that proved to be glandular/dense tissue.
4. True Negative (TN) means breast segmented or classified as other tissues that proved to be other tissues.

$$TPF = \frac{TP}{TP+FN} \qquad (1)$$

$$FPF = \frac{FP}{FP+TN} \qquad (2)$$

There are researchers that evaluate the performance of segmentation results using 2 performance metrics: completeness (CM) and correctness (CR) [26, 60]. Completeness is the percentage of the ground truth region which is explained by the segmented region. Correctness is the percentage of correctly extracted breast region type. A single metric which is quality, can be obtained by combining completeness and correctness [26, 46]. The optimum value for both metrics is 1.

$$Completeness = \frac{TP}{TP+FN} \qquad (3)$$

$$Correctness = \frac{FP}{FP+TN} \qquad (4)$$

$$Quality = \frac{TP}{TP+FN+FP} \qquad (5)$$

However, the problem here is that the qualitative response of the radiologist is very subjective and varies hugely [55, 58, 59]. The ground truth by each radiologist may be different from one radiologist to another. Each researcher would try to obtain the ground truth from the radiologist and compared the performance of their research segmentation result with other researchers. According to Nishikawa *et al.* [61], it is not meaningful to compare different techniques if the techniques are tested on different databases. Even so, the problem is, sometimes the same database were used but with different ground truths. So, how do we measure the reliability of the performance of the segmentation result? It is necessary to find a way to obtain the objective ground truth.

According to Olsen and Georgsson, it is very difficult to obtain the objective ground truth [62]. They have proposed a method to relate markings of the ground truth between groups of radiologists to achieve levels of agreement. Consequently, the problem which might arise was that many ground truths need to be taken and this proves to be time consuming unless it involves only a small amount of data. Markings for ground truth depend on hands-on capability and skill. For example, radiologist who is very careful, meticulous and experienced can give more detailed ground truth markings distinguishing ducts and lobules. On the other hand, a radiologist who is not too diligent may give a rough outline by inserting the whole glandular region. This practice may give rise to the inclusion of the fatty regions in the area of interest. There are researchers who try to propose their own performance measurement methods [60]. However, the accuracy in these could be disputed because their studies were based on their own ground truth and comparisons were made with another research, with different ground truths. This makes it impossible for the measurement accurately comparable.

## 5. Recommendation

Classification of glandular tissue is beneficial for estimation of breast density for categorizing it and also to establish an optimal strategy to follow if there is suspicious region, while segmentation of glandular tissue can visualize the suspicious region. Furthermore, segmentation of breast anatomical region can give more specific delineation of breast tissue to help radiologist in the interpretation. Therefore, for future work, it is important to combine segmentation of the breast into anatomical regions with the segmentation of glandular tissue for general breast cancer screening. Then, focusing on the dense component, specific segmentations of glandular tissue areas should be adapted for breast lesion characterizations. Finally, breast density estimation for breast cancer risk assessment or for monitoring the changes in breast density as prevention or intervention procedure, should also be incorporated. Therefore, future works should combine all the steps in the Computer Aided Diagnosis System.

In performance evaluation, there is still no standard measurement or an objective ground truth for the mammogram image that had been segmented as yet. Hence, future research should try to identify the same ground truth in order to compare the computer assisted system that will be developed.

### Acknowledgments

The authors would like to acknowledge USM-RU Grant 814082 for providing financial support for this work.

### References

[1] M. J. Yaffe, "Mammographic density: Measurement of mammographic density", Breast Cancer Research, Vol. 10, No. 3, 2008.
[2] K. E. Martin, M. A. Helvie, C. Zhou, M. A. Roubidoux, J. E. Bailey, C. Paramagul, C. E. Blane, K. A. Klein, S. S. Sonnad, H-P Chan, "Mammographic Density Measured with Quantitative Computer-aided Method- Comparison with Radiologists' Estimates and BI-RADS Categories", Radiology, vol. 240, 2006, pp. 656-665.
[3] J. J. Heine, P. Malhotra, "Mammographic tissue, breast cancer risk, serial image analysis, and digital mammography", Acad. Radiol., Vol.9, 2002, pp. 298–335.
[4] P. C. Stomper, B. J. Van Voorhis, V. A. Ravnikar, J. E. Meyer, "Mammographic changes associated with postmenopausal hormone replacement therapy: a longitudinal study", Radiology, Vol. 174, 1990, pp. 487–490.
[5] M. B. Laya, J. C. Gallagher, J. S. Schreiman, E. B. Larson, P. Watson, L. Weinstein, "Effect of postmenopausal hormonal replacement therapy on mammographic density and parenchymal pattern", Radiology, Vol. 196, 1995, pp. 433–437.
[6] H. J. Son, K. K. Oh, "Significance of follow-up mammography in estimating the effect of tamoxifen in breast cancer patients who have undergone surgery", AJR Am J Roentgenol, Vol. 173, 1999, pp. 905–909.
[7] G. A. Colditz, S. E. Hankinson, D. J. Hunter, W. C. Willett, J. E. Manson, M. J. Stampfer, C. Hennekens, B. Rosner, F. E. Speizer, "The use of estrogens and progestins and the risk of breast cancer in postmenopausal women", N Engl J Med, Vol. 332, 1995, pp. 1589–1593.
[8] R. K. Ross, A. Paganini-Hill, P. C. Wan, M. C. Pike, "Effect of hormone replacement therapy on breast cancer risk: estrogen versus estrogen plus progestin", J Natl Cancer Inst, Vol. 92, 2000, pp. 328–332.
[9] B. Fisher, J. P. Costantino, D. L. Wickerham, C. K. Redmond, M. Kavanah, W. M. Cronin, V. Vogel, A. Robidoux, N. Dimitrov, J. Atkins, M. Daly, S. Wieand, E. Tan-Chiu, L. Ford, N. Wolmark, "Tamoxifen for prevention of breast cancer: report of the National Surgical Adjuvant Breast and Bowel Project P-1 Study", J Natl Cancer Inst, Vol. 90, 1998, pp. 1371–1388.
[10] J. E. Rossouw, G. L. Anderson, R. L. Prentice, *et al.* "Risks and benefits of estrogen plus progestin in healthy postmenopausal women: principal results from the Women's Health Initiative randomized controlled trial", *JAMA*, Vol. 288, 2002, pp. 321–333.
[11] S. Buseman, J. Mouchawar, N. Calonge, T. Byers, "Mammography screening matters for young women with breast carcinoma", Cancer, Vol. 97, 2003, pp. 352-358.
[12] GLOBOCAN 2008, "Cancer Incidence and Mortality Worldwide in 2008," The International Agency for Research on Cancer (IARC). Available: http://globocan.iarc.fr/
[13] A. N. Hisham, C.-H. Yip, "Overview of Breast Cancer in Malaysian Women: A Problem with Late Diagnosis", Asian Journal of Surgery, Vol. 27, no. 2, 2004, pp. 130-133.
[14] T. S. Subashini, V. Ramalingam, S. Palanivel, "Automated assessment of breast tissue density in digital mammograms," Computer Vision and Image Understanding, Vol. 114, No. 1, 2010, pp. 33–43.
[15] G. Agarwal, P. V. Pradeep, V. Aggarwal, C-H. Yip and P. S. Y. Cheung, "Spectrum of Breast Cancer in Asian Women", World Journal of Surgery, Vol. 31, No. 5, 2007, pp. 1031-1040.
[16] M. Pierre, "Combining assembles of domain expert markings", M. Sc. Thesis, Department of Computing Science, Umeåa University, Sweden, May 30, 2010.
[17] J. N. Wolfe, "Breast patterns as an index of risk for developing breast cancer", Journal of Roentgenology, Vol. 26, 1976, pp. 1130–1139.
[18] N. F. Boyd, J. M. Rommens, K. Vogt, V. Lee, J. L. Hopper, M. J. Yaffe, A. D. Paterson, "Mammographic breast density as an intermediate phenotype for breast cancer", Lancet Oncology, Vol. 6, 2005, pp. 798–808.
[19] E. J. Aiello, D. S. Buist, E. White, P. L. Porter, "Association between mammographic breast density and breast cancer tumor characteristics", Cancer Epidemiology Biomarkers and Prevention, Vol. 14, 2005, pp. 662–668.
[20] L. A. Habel, J. J. Dignam, S. R. Land, "Mammographic density and breast cancer after ductal carcinoma in situ", Journal of the National Cancer Institute, Vol. 96, No. 19, 2004, pp. 1467–1472.

[21] M. L. Irwin, E. J. Aiello, A. McTiernan, L. Bernstein, F. D. Gilliland, R. N. Baumgartner, K. B. Baumgartner, R. Ballard-Barbash, "Physical activity, body mass index, and mammographic density in postmenopausal breast cancer survivors", Journal of Clinical Oncology, Vol. 25, No. 9, 2007, pp. 1061–1066.

[22] J. N. Wolfe, "Risk for breast cancer development determined by mammographic parenchymal pattern", Cancer, Vol. 37, 1976, pp. 2486–2492.

[23] N. F. Boyd, J. W. Byng, R. A. Long, E. K. Fishell, L. E. Little, A. B. Miller, G. A. Lockwood, D. L. Tritchler and M. J. Yaffe, "Quantitative classification of mammographic densities and breast cancer risk: results from the Canadian National Breast Screening study", J. Nat. Cancer Inst., Vol. 87, 1995, pp. 670–675.

[24] C. H. van Gils, J. H. Hendriks, R. Holland, N. Karssemeijer, J. D. Otten, H. Straatman, A. L. Verbeek, "Changes in mammographic breast density and concomitant changes in breast cancer risk", Eur J Cancer Prev., Vol. 8, No. 6, 1999, pp. 509-515.

[25] N. Karssemeijer, "Automated classification of parenchymal patterns in mammograms", Phys. Med. Biol., Vol. 43, 1998, pp. 365–378.

[26] W. Wirth, D. Nikitenko, J. Lyon, "Segmentation of Breast Region in Mammograms using a Rule-Based Fuzzy Reasoning Algorithm", ICGST Graphics, Vision and Image Processing Journal, Vol. 5, No. 2, 2005, pp. 45-54.

[27] L. W. Bassett and R. H. Gold, Breast Cancer Detection: Mammography and Other Methods in Breast Imaging, 2nd edition, Grune & Stratton, Orlando, FL, 1987.

[28] S. Caulkin, S. Astley, J. Asquith and C. Boggis "Sites of occurrence of malignancies in mammograms," In N. Karssemeijer, M. Thijssen, J. Hendriks and L. Van Erning, Proceedings of the 4th International Workshop on Digital Mammography, Nijmegen, The Netherlands, June, 1998, pp. 279-282.

[29] J. W. Byng, N. F. Boyd, E. Fishell, R. A. Jong, M. J.Yaffe, "Automated analysis of mammographic densities", Phys. Med. Biol., Vol. 41, 1996, pp. 909–923.

[30] American College of Radiology, American College of Radiology Breast Imaging Reporting and Data System (BIRADS). 4th ed., American College of Radiology, Reston, VA, 2003.

[31] C. Zhou, H. P. Chan, N. Petrick, M. A. Helvie, M. M. Goodsitt, B. Sahiner, and L. M. Hadjiiski, "Computerized image analysis: estimation of breast density on mammograms", Medical Physics, Vol. 28, No. 6, 2001, pp. 1056–1069.

[32] J. Suckling, D. R. Dance, E. Moskovic, D. J. Lewis and S. G. Blacker, "Segmentation of mammograms using multiple linked self-organizing neural networks", Med. Phys., Vol. 22, No. 2, 1995, pp. 145–52.

[33] P. Miller and S. M. Astley, "Classification of breast tissue by texture analysis", Image Vision Comput., Vol. 10, 1992, pp. 277–282.

[34] T. Matsubara, D. Yamazaki, M. Kato, T. Hara, H. Fujita, T. Iwase, T.Endo, "An automated classification scheme for mammograms based on amount and distribution of fibroglandular breast tissue density", International Congress, series 1230, 2001.

[35] R. J. Ferrari, R. M. Rangayyan, R. A. Borges, and A. F. Frere, "Segmentation of the fibro-glandular disc in mammograms via Gaussian mixture modeling", Med. Biol. Eng. Comput, 2004, Vol. 42, pp.378–387.

[36] C. Ols´en and A. Mukhdoomi, "Automatic Segmentation of Fibroglandular Tissue", LNCS 4522. In: B.K. Ersbøll and K.S. Pedersen (Eds.): SCIA 2007, 2007.

[37] M. Masek, "Hierarchical Segmentation of Mammograms Based on Pixel Intensity", PhD thesis, Centre for Intelligent Information Processing Systems, School of Electrical, Electronic, and Computer Engineering. University of Western Australia, Crawley, WA, February, 2004.

[38] A. El-Zaart, "Expectation–maximization technique for fibro-glandular discs detection in mammography images", Comput Biol Med., Vol. 40, No. 4, 2010, pp. 392-401,

[39] P. Taylor, S. Hajnal, M-H Dilhuydy, and B. Barreau, "Measuring image texture to separate difficult from easy mammograms", The British Journal of Radiology, Vol. 67, 1994, pp. 456–463.

[40] K. Bovis and S. Singh, "Classification of mammographic breast density using a combined classifier paradigm", in Proc. Med. Image Understanding Anal. Conf., 2002, pp. 177–180.

[41] J Kittler, M. Hatef, R. P. W Duin, and J. Matas, "On combining classifiers," IEEE Transactions on Pattern Analysis and Machine Intelligence, Vol. 20, No. 3, 1998, pp. 226–239.

[42] A. Torrent, A. Bardera, A. Oliver, J. Freixenet, I. Boada, M. Feix and J. Martı́, "Breast Density Segmentation: A Comparison of Clustering and Region Based Techniques", (eds.) IWDM 2008. LNCS, Springer, Heidelberg, Vol. 5116, 2008, pp. 9–16.

[43] A. Oliver, J. Freixenet, R. Martí, J. Pont, E. Pérez, E. R. Denton, R. Zwiggelaar, "A Novel Breast Tissue Density Classification Methodology", IEEE Trans Inf Technol Biomed, Vol. 12, No. 1, 2 008,pp. 55-65.

[44] A. Oliver, X. Lladó, E. Pérez, J. Pont, E. R. E. Denton, J. Freixenet, and J. Martí, "A Statistical Approach for Breast Density Segmentation", Journal of Digital Imaging, 2009, pp. 1-11.

[45] N. Saidin, U. K. Ngah, H. A. M. Sakim, D. N. Siong and M. K. Hoe, "Density Based Breast Segmentation for Mammograms Using Graph Cut Techniques", in TENCON (IEEE Region 10 Conf.), 2009, pp. 1-5.

[46] M. Adel, M. Rasigni, S. Bourennane, and V. Juhan, "Statistical segmentation of regions of interest on a mammographic image", EURASIP Journal on Advances in Signal Processing, Vol. 2007, Article ID 49482, 2007, pp. 1-8.

[47] S. R. Aylward, B. M. Hemminger, E. D. Pisano, "Mixture Modeling for Digital Mammogram Display and Analysis", in *Digital Mammography*, N. Karssemeijer, M. A. O. Thijssen, J. H. C. L. Hendriks, L. J. T. O. Van Erning, editors, Computational Imaging and Vision Series, Vol. 13, Kluwer Academic Publishers, Dordrecht, 1998, pp. 305-312.

[48] R. D. Yapa, K. Harada, "Breast Skin-Line Estimation and Breast Segmentation in Mammograms using Fast-Matching Method", Int. J. of Biological and Medical Sciences, Vol. 3, no. 1, 2008, pp. 54-62.

[49] S.Chatzistergos, J. Stoitsis, K. S. Nikita, A. Papaevangelou, "Development of an integrated breast tissue density classification software system", *IEEE International Workshop on Imaging Systems and Technique*s (IST 2008), 10-12 Sept. 2008, pp. 243 – 245.
[50] A. Bosch, X. Munoz, A. Oliver, J. Marti, "Modeling and Classifying Breast Tissue Density in Mammograms", IEEE Computer Society Conference on Computer Vision and Pattern Recognition (CVPR'06), vol. 2, 2006,pp. 1552 – 1558.
[51] T. Hofmann, "Unsupervised learning by probabilistic latent semantic analysis," Machine Learning, Vol. 41, No.2, 2001, pp.177-196.
[52] M. Heath, K. Bowyer, D. Kopans, R. Moore and P. Kegelmeyer Jr, "The digital database for screening mammography", Proceedings of the Fifth International Workshop on Digital Mammography, 2001, pp 212–218.
[53] J. Suckling, et al: "The mammographic image analysis society digital mammogram database", *Exerpta Medica International Congress,* series 1069, pp. 375–378, 1994.
[54] S. R. Amendolia, et al: "The CALMA project," Nuclear Instruments & Methods in Physics Research Section A: Accelerators, Spectrometers, Detectors & Assoc. Equipment," Vol. 461, issues 1-3, pp. 428–429, 2001.
[55] Lawrence Livermore National Library/UCSF Digital Mammogram Database. Center for Health Care Technologies Livermore. Livermore, CA, USA.
[56] B. R. N. Matheus & H. Schiabel. Online Mammographic Images Database for Development and Comparison of CAD Schemes. *Journal of Digital Imaging*, pp. 1-7, 2010.
[57] M. A. Wirth, "Performance Evaluation of CADe Algorithms in Mammography", in Recent Advances in Breast Imaging, Mammography, and Computer-Aided Diagnosis of Breast Cancer, Jasjit S. Suri, Rangaraj M. Rangayyan, editors, SPIE Press, Bellingham, WA, pp. 640-671.
[58] C. E. Metz, "Evaluation of digital mammography by ROC analysis," In Proc. International Workshop on Digital Mammography, pp. 61–68, 1996.
[59] D. P. Chakraborty, H. J. Yoona, and C. Mello-Thoms, "Localization accuracy of radiologists in free-response studies: Inferring perceptual FROC curves from mark-rating data," Academic Radiology, Vol. 14, 2007, pp.4–18.
[60] M. Wirth, J. Lyon, M. Fraschini, D. Nikitenko, "The effect of mammogram databases on algorithm performance", in Proceedings of the 17th IEEE Symposium on Computer-Based Medical Systems (CBMS'04), 2004.
[61] R. H. Nishikawa, M. L. Giger, K. Doi, C. E. Metz, F. F.Yin, C. J. Vyborny, R. A. Schmidt, "Effect of case selection on the performance of computer-aided detection schemes", Medical Physics, AAPM, 1994, pp.265-269.
[62] C. Ols´en and F. Georgsson, "Assessing Ground Truth of Glandular Tissue," In: Susan M. Astley et al. (Eds.): IWDM 2006, LNCS 4046.

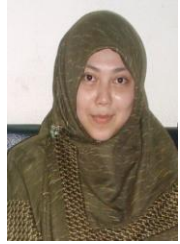

**Nafiza Saidin** received her B.Eng degree (Medical Electronic) from Universiti Teknologi Malaysia in August, 1999. She subsequently undertook research at Universiti Sains Malaysia, honoured her with an M.Sc. degree (Medical Imaging) in 2005. She has published a number of conference papers, including a book chapter and her main interests are in biomedical engineering, image processing and medical imaging. Currently, she is a postgraduate student at PhD level at Universiti Sains Malaysia. She is a member of IEEE.

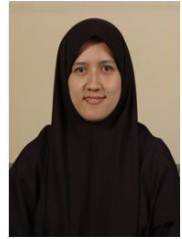

**Harsa Amylia Mat Sakim** received the Bachelor of Engineering degree from the University of Liverpool and the M.Sc. from University of Newcastle Upon Tyne, UK. She obtained her PhD from School of Electrical and Electronic Engineering at Universiti Sains Malaysia, where she is now teaching and pursuing her passion in research. She has published papers in international journals, specifically in breast cancer studies. Her research interests include Artificial Intelligence, Biomedical Engineering and Medical Electronics and Instrumentation. She is a member of IEEE.

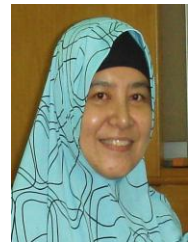

**Associate Professor Dr. Umi Kalthum Ngah**, (B.Sc. (Hons) Sheffield, M.Sc. (USM), Ph.D (USM)) received her B.Sc. (Hons.) in Computer Science from the University of Sheffield in 1981. In 1995, she received her M.Sc. in Electronic Engineering (majoring in Image Processing and Knowledge Based Systems) from Universiti Sains Malaysia and then pursued further degree at the same university where she received her PhD in the same area in the year 2007. She has been with USM since the year 1981, starting her career as a tutor. Currently, she is attached to the School of Electrical and Electronic Engineering, USM Engineering Campus. Her current research interests include image processing, particularly medical imaging, knowledge based and artificial intelligence systems (including animal inspired optimization techniques) and biomedical engineering focusing on intelligent systems. Her work has been published in numerous international and national journals, chapters in books, international and national proceedings.

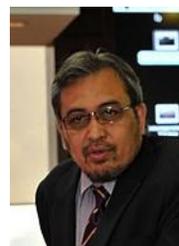

**Professor Dr. Ibrahim Lutfi Shuaib**, (MBBS (UM), DMRD (Liverpool), FRCR (UK)) graduated from University of Malaya, Kuala Lumpur in 1985 with MBBS. In 1988, he joined the Merseyside radiology training scheme in Liverpool, UK as an honorary registrar. He obtained Diploma in Diagnostic Radiology (DMRD) from University of Liverpool in 1990. Following that, he joined Leicestershire radiology training scheme in Leicester, UK as a registrar. In 1991, he was accepted to join the radiology training scheme as a senior registrar in the Merseyside, Liverpool area. He returned to Universiti Sains Malaysia, Health Campus, Kubang Kerian in 1993 as a lecturer after obtaining FRCR (UK) in 1992. He is now working in Advanced Medical and Dental Institute, Kepala Batas. His interest is in musculoskeletal radiology and Health Informatics.